\documentclass[letterpaper]{article} 
\usepackage{aaai24}  
\usepackage{times}  
\usepackage{helvet}  
\usepackage{courier}  
\usepackage[hyphens]{url}  
\usepackage{graphicx} 
\usepackage{natbib}  
\usepackage{caption} 
\usepackage{algorithm}
\usepackage{algorithmic}
\usepackage{bbding}
\usepackage{graphicx}
\usepackage{multirow}
\usepackage{dutchcal}
\usepackage{color}

\usepackage{newfloat}
\usepackage{listings}
\DeclareCaptionStyle{ruled}{labelfont=normalfont,labelsep=colon,strut=off} 
\floatstyle{ruled}
\newfloat{listing}{tb}{lst}{}
\floatname{listing}{Listing}
\title{M-BEV: Masked BEV Perception for Robust Autonomous Driving}
\author{
    Siran Chen\textsuperscript{\rm 1,\rm 2},
    Yue Ma \textsuperscript{\rm 4},
    Yu Qiao\textsuperscript{\rm 1, \rm 3},
    Yali Wang\textsuperscript{\rm 1, \rm3,} \thanks{Corresponding author}
}

\affiliations{
    \textsuperscript{\rm 1} Shenzhen Institute of Advanced Technology, Chinese Academy of Science, Shenzhen, China\\
    \textsuperscript{\rm 2} School of Artificial Intelligence, University of Chinese Academy of Science, Beijing, China \\
    \textsuperscript{\rm 3} Shanghai Artificial Intelligence Laboratory, Shanghai, China \\
    \textsuperscript{\rm 4} Tsinghua Shenzhen International Graduate School, Tsinghua University, Shenzhen, China\\
    Chensiran17@mails.ucas.ac.cn, y-ma21@mails.Tsinghua.edu.cn, \\ 
    \{yu.qiao, yl.wang\}@siat.ac.cn
}

\usepackage{bibentry}

\begin{document}

\maketitle

\begin{abstract}
3D perception is a critical problem in autonomous driving.
Recently,
the Bird’s-Eye-View (BEV) approach has attracted extensive attention,
due to low-cost deployment and desirable vision detection capacity.
However,
the existing models ignore a realistic scenario during the driving procedure,
i.e.,
one or more view cameras may be failed,
which largely deteriorates the performance. 
To tackle this problem,
we propose a generic Masked BEV (M-BEV) perception framework,
which can effectively improve robustness to this challenging scenario,
by random masking and reconstructing camera views in the end-to-end training.
More specifically,
we develop a novel Masked View Reconstruction (MVR) module for M-BEV.
It mimics various missing cases by randomly masking features of different camera views,
then leverages the original features of these views as self-supervision,
and reconstructs the masked ones with the distinct spatio-temporal context across views.
Via such a plug-and-play MVR,
our M-BEV is capable of learning the missing views from the resting ones,
and thus well generalized for robust view recovery and accurate perception in the testing.
We perform extensive experiments on the popular NuScenes benchmark,
where
our framework can significantly boost 3D perception performance of the state-of-the-art models on various missing view cases,
e.g., 
for the absence of back view, 
our M-BEV promotes the PETRv2 model with 10.3\% mAP gain.

\end{abstract}

\begin{figure}
   \centering
   \includegraphics[width=0.49\textwidth]{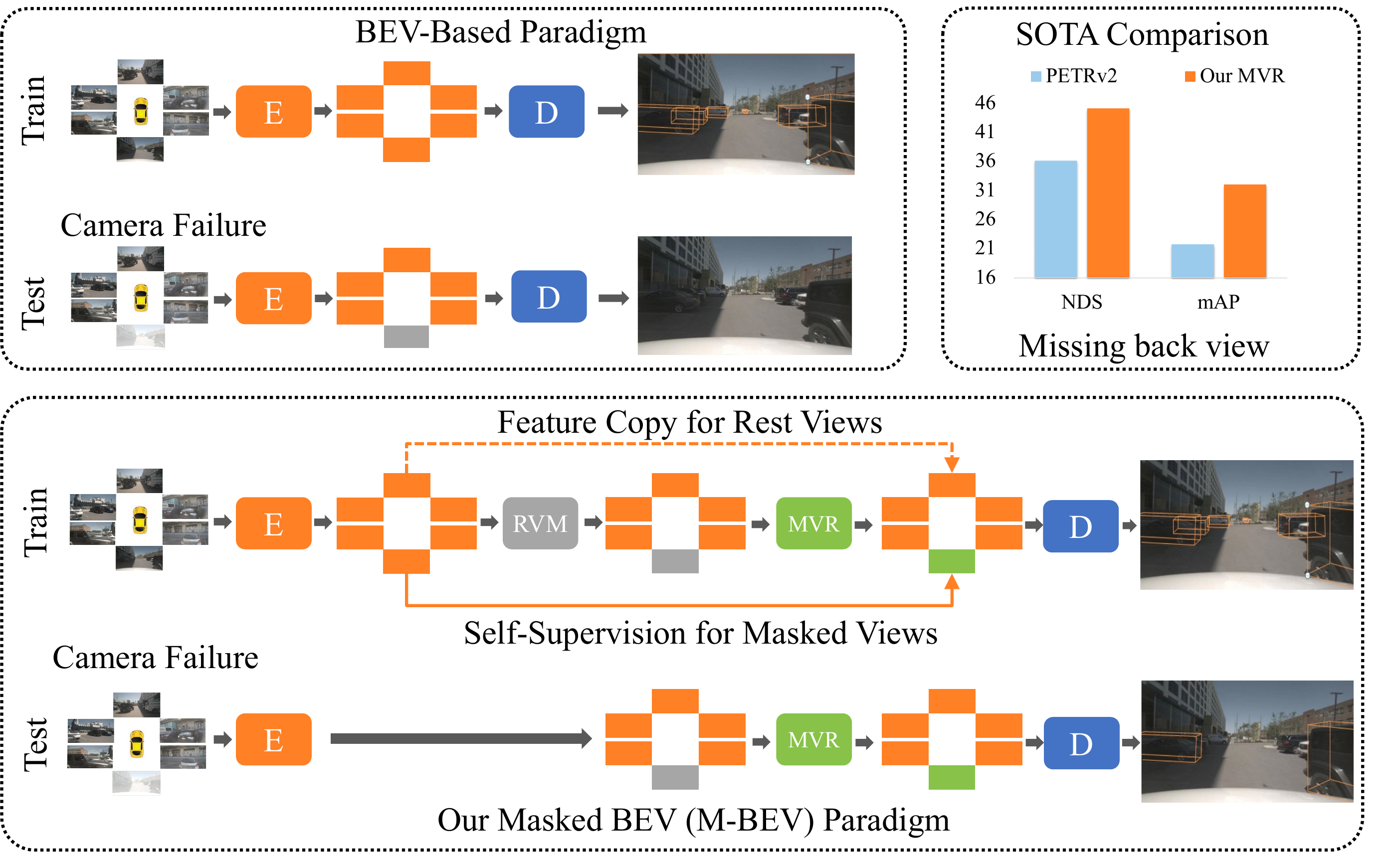}
   \caption{Motivation. In cases where cameras crash, the existing BEV-based approaches would be largely deteriorated. We design a self-supervised Masked View Reconstruction (MVR) module which can significantly boost the state-of-the-art models for various missing camera cases. E: Encoder, D: Decoder, RVM: Random View Masking.
   }
   \label{fig:teaser}
\vspace{-0.5cm}     
\end{figure}

\section{Introduction}
3D perception of surrounding scenes is the key for autonomous driving.
Compared to LiDAR-based methods \cite{mohapatra2021bevdetnet,barrera2020birdnet+,ma2022cg,zhou2020end},
camera-based approaches has attracted increasing attention\cite{li2022bevformer,philion2020lift,roddick2018orthographic,wang2019pseudo} 
since they are easy and cheap for deployment.
In particular,
the Bird’s-Eye-View (BEV) based methods have been highlighted by learning the holistic representation from multi-camera images \cite{roddick2018orthographic,wang2019pseudo,philion2020lift,li2022bevformer,li2022bevdepth,li2023bevstereo,wang2022detr3d, liu2022petr, chen2023attentive}.
Basically,
these approaches integrate 2D image information from six distinct views to encode a unified 3D representation of visible scenes, 
and then decode it to accurately capture the size and location of objects in the surrounding. 
However,
these approaches work on the ideal case in which six cameras always work well,
while
one or more cameras may be failed or broken down during the realistic driving procedure.
In such an emergency,
the existing BEV-based approaches would be largely deteriorated,
due to the lack of the corresponding visual clues from the missing views.
For example, 
the NDS and mAP of PETRv2 \cite{liu2023petrv2} have a decrease of 12.4\% and 18.0\% respectively when the back camera view is missing, which severely affects the safety and reliability of autonomous driving system. 

To alleviate this problem,
we propose a concise Masked BEV (M-BEV) perception framework, 
which can effectively boost model's robustness to missing camera views,
by randomly masking and recovering view features in the end-to-end training procedure.
Specifically,
we design a self-supervised Masked View Reconstruction (MVR) module in our M-BEV,
inspired by the style of the well-known Masked Autoencoders (MAE) \cite{he2022masked}.
In particular,
we randomly mask the features of different camera views in the training epochs.
Then,
we leverage the features of the rest views as spatio-temporal context,
and recover the features of the masked views as their original features.
Via such learning,
MVR masters the capacity of reconstructing the missing views from the rest views,
and thus can effectively tackle the camera failure cases in the testing.

Note that,
there are two critical differences between our MVR and MAE ~\cite{he2022masked}.
\textit{First},
the masking goal and design are different.
MAE aims at learning scalable image representation in general.
Hence,
it masks several \textit{patches} of each image and reconstructs them from the rest patches.
Alternatively,
our MVR aims at tackling the missing camera cases in BEV-based driving.
Hence,
it masks several \textit{images} of six camera views and reconstructs them from the rest images.
More importantly,
besides using all the rest images for reconstruction,
we also propose to exploit the distinct contexts in the surrounding images for reconstruction,
based on spatio-temporal overlaps across BEV cameras.
MAE does not take these thoughts into account.
\textit{Second},
the training and testing manners are different.
In the training stage,
MAE uses the \textit{same} proportion for masking patches of each input image in all the epochs,
in order to train large image models in general.
Alternatively,
our MVR uses the \textit{different} proportion for masking images of six views in the different epochs,
in order to contain various missing camera views which can possibly happen.
In the testing stage,
MAE mainly uses \textit{encoder} as general feature extractor and ignores the decoder.
Alternatively,
our MVR uses \textit{decoder} to reconstruct the missing camera views and leverages the recovered views for 3D perception in the BEV-based driving.

Finally,
we implement our general M-BEV framework on two state-of-the-art BEV-based models,
i.e.,
PETRv2~\cite{liu2023petrv2}
,
and 
BEVStereo \cite{li2023bevstereo}
for 3D object detection in autonomous driving.
Note that,
we choose these models for evaluation,
since they maintain a preferable accuracy-efficiency balance and we have limited computation resources.
To verify the effectiveness,
we perform extensive experiments on various missing camera cases,
on the popular NuScenes\cite{caesar2020nuscenes} dataset.
The results show that,
M-BEV framework significantly boosts the performance of the SOTA models for various missing camera emergencies.
For example, for the absence of back view, M-BEV helps to boost PETRv2 baseline with 10.3\% mAP improvement, while only takes extra 0.6ms for once response.


\section{Related Work}


\noindent\textbf{Multi-View 3D Object Detection}.
3D object detection is one of the key technologies for autonomous driving, 
which takes 
LiDAR \cite{zhou2018voxelnet,ma2022simvtp, li2021sienet,lang2019pointpillars}, 
camera \cite{philion2020lift,huang2021bevdet,liu2022petr, wang2021fcos3d},
or multi-modal input data \cite{liu2022bevfusion,xu2021fusionpainting, ma2022visual, yin2021multimodal}
to predict the location, size, velocity, and category of the targets in real 3D space. 
Cameras-based methods \cite{li2022bevformer,liu2023petrv2,ma2023follow, li2022bevdepth,li2023bevstereo} stand out,
due to the low cost and easy access to visual data from six camera views.
The key problem of these works is the conversion between 2D and 3D space,
and BEV representation works as a suitable bond.
OFT\cite{roddick2018orthographic} first makes a direct transformation from 2D features to 3D BEV features for monocular 3D object detection.
The following works expand this style,
by learnable 3D object queries~\cite{wang2022detr3d},
3D position-aware embedding \cite{liu2022petr,liu2023petrv2},
temporal information integration \cite{li2022bevformer,ma2023magicstick, huang2021bevdet,huang2022bevdet4d,liu2023petrv2}, etc.
Additionally,
depth supervision is another main direction for 3D performance enhancement 
\cite{li2022bevdepth,chen2022bevdistill,huang2022tig,li2023bevstereo}.
However,
all of these existing BEV-based methods rely on the high-quality camera inputs in the ideal case.
When the camera views fails in practice,
their performance declines severely.
RoboBEV~\cite{zhu2023understanding} establishes a comprehensive driving benchmark under various natural and adversarial corruptions.
MetaBEV\cite{ge2023metabev} solves the problem by cross-modal data fusion with both Camera and LiDAR.
As far as we know, 
our M-BEV is the first camera-only solution for such view failure with great robustness.

\noindent\textbf{Masked Visual Modeling}.
Masked modeling pipeline was first used in NLP \cite{radford2019language}. 
The masking operation is treated as a noise type and processed by traditional denoising encoders\cite{vincent2008extracting}. 
Then ViT\cite{dosovitskiy2020image} uses masked token prediction and paves the way for self-supervised pre-training.
More recently, 
MAE\cite{he2022masked} is introduced as an asymmetric transformer-based encoder-decoder architecture, 
which is achieved by reconstructing the pixels of the masked image.
The pre-trained autoencoder could be applied for various downstream tasks with minor fine-tuning. 
Similar thoughts are raised by BEiT\cite{bao2021beit}, BEVT\cite{wang2022bevt}, and VIMPAC\cite{tan2021vimpac}  while the reconstruction is based on token-level. 
MaskFeat\cite{wei2022masked} chooses to reconstruct HOG\cite{dalal2005histograms} features of the masked token as self-supervised pre-training.
Moreover,
UM-MAE\cite{li2022uniform} and SemMAE\cite{li2022semmae} design distinctive masking strategies, 
VideoMAE\cite{tong2022videomae, wang2023videomae} series simply expand masking in the temporal dimension of videos and achieve impressive performance. 
Compared to MAE-style design,
our MVR has two critical differences which has been carefully discussed in the introduction.

\begin{figure}[t]
\centering
\includegraphics[width=0.49\textwidth]{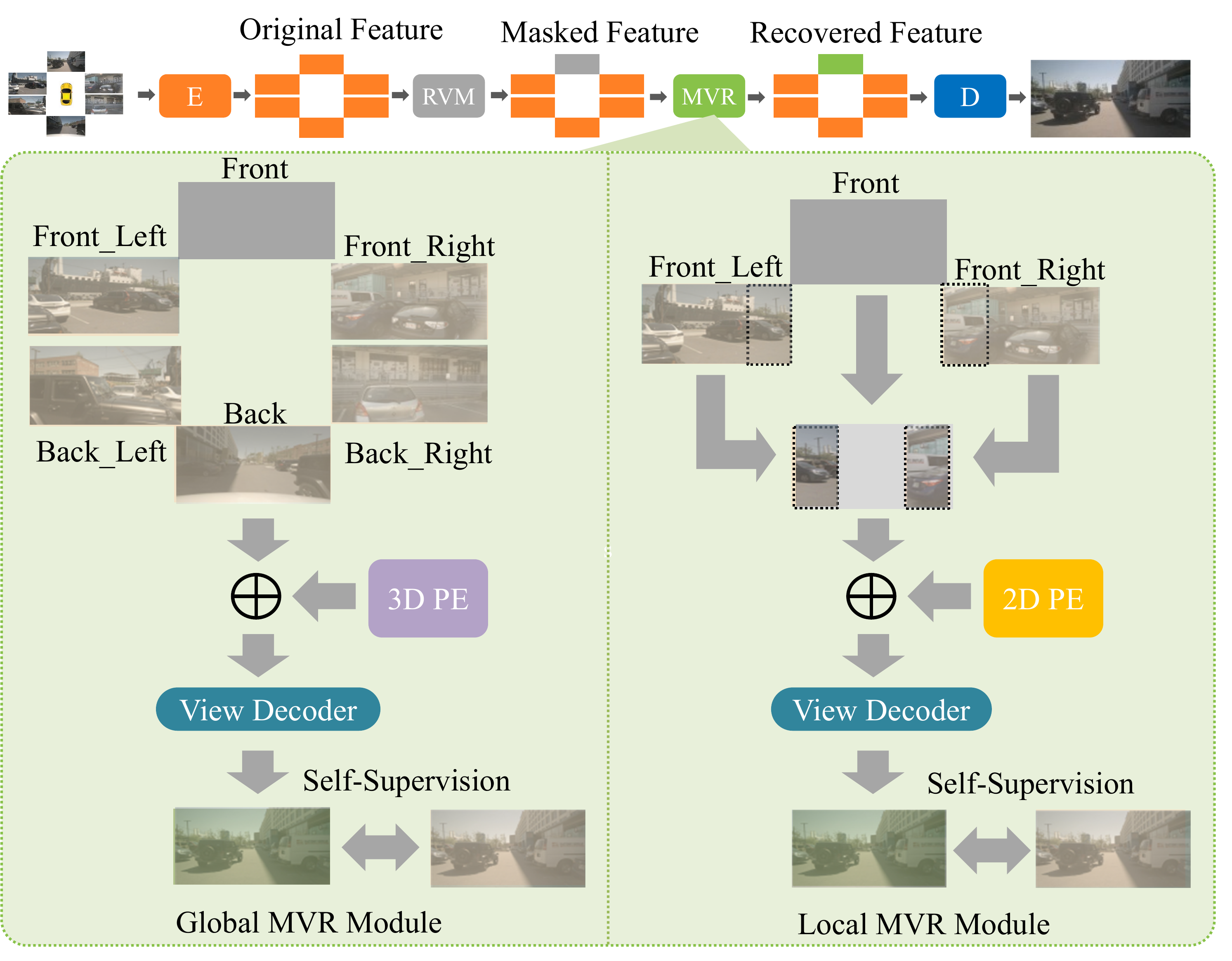}
    \caption{Overall Architecture of Our Proposed M-BEV Framework. M-BEV consists of 3D Object Detector with Visual Encoder (E) and Decoder (D), Random View Masking (RVM) and Masked View Reconstruction (MVR) modules. We propose two MVR modules, namely Global MVR and Local MVR,
    to recover the features of masked views. }
\label{fig:Overall Architecture}
\vspace{-0.5cm}     
\end{figure}

\section{Method}
To deal with the camera failure,
we introduce our Masked BEV paradigm (M-BEV) in this section.
First,
we introduce the overall model architecture of M-BEV.
Then,
we have a detailed description of the critical Masked View Reconstruction (MVR) module,
and explain how it works to boost the robustness of camera failure.
Finally,
we illustrate the training and testing procedure of M-BEV.

\subsection{Overall Architecture}
\label{sec:overarc}

The overall architecture of M-BEV is shown in Fig. \ref{fig:Overall Architecture},
it consists of a 3D object detector, Random View Masking (RVM), and Masked View Reconstruction (MVR) modules. 
Basically,
the 3D object detector contains a visual encoder, translator, and decoder.
First,
the encoder transforms the input images of multiple camera views into their corresponding 2D visual features.
Second,
the translator transforms these 2D features into 3D-relevant features,
by 
3D position embedding \cite{liu2023petrv2,liu2022petr}, 
depth estimation \cite{li2022bevdepth, li2023bevstereo, chen2022bevdistill},
attention mechanism \cite{li2022bevformer, yang2022bevformer, ma2022cmal, wang2022detr3d},
etc. 
Finally,
the decoder transforms these 3D features for final object detection.
Note that,
when one or more cameras fail,
there are no corresponding features from encoder.
To address this, we incorporate the RVM and MVR modules after the encoder to recover the missing features. 
In this case,
we denote the combination of translator and decoder as an integrated decoder in this paper without loss of generality.

To mimic the emergency situation of camera failure in the testing phrase,
we propose to perform view masking and recovering in the training phrase.
Specifically,
RVM is used to randomly select the camera views for masking,
and
MVR is used to recover the masked views by exploiting spatio-temporal context from the rest views.
Note that,
recovering is performed on the feature-level of the camera views from encoder,
instead of raw images like in MAE \cite{he2022masked}.
The main reason is that
our goal is to reconstruct the missing views for 3D object detection,
if we reconstruct the raw images,
these predicted images have to be fed into the visual encoder again to obtain their features for detection, this makes the whole paradigm tedious with unnecessary processing and computation.
Alternatively,
if we reconstruct the features of the masked images,
we could directly use the visual encoder(e.g., VoVNet, ViT, ResNet, etc) in the BEV-based models for both detection and reconstruction.
These recovered features can be straightforwardly used for subsequent decoding without any difficulty. 
In this case,
we can effectively leverage the self-supervised advantages of mask modeling,
by only adding MVR with a lightweight feature decoder.
In the following,
we will explain how to design MVR in detail for reconstructing the masked views.

\subsection{Masked View Reconstruction Module}
\label{sec:masked_strategy}

\begin{figure}
\begin{center}
\setlength{\abovecaptionskip}{0.cm}
\includegraphics[width=\columnwidth]{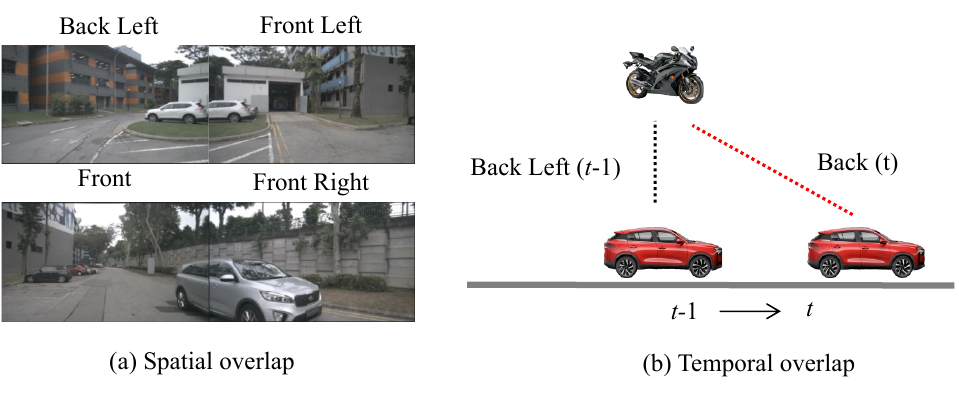}    
\end{center}
    \caption{Illustration of spatial and temporal overlap. }
\label{fig:view}
\vspace{-0.3cm}     
\end{figure}

Typically,
BEV-based approaches ~\cite{li2022bevformer, huang2022bevdet4d, liu2023petrv2, jiang2022polarformer, li2022bevdepth, lin2022sparse4d} take the images of six camera views at previous step $t-1$ and current step $t$ as input.
After encoding these images as visual features,
we use RVM module to randomly mask the features of several camera views.
Note that in real situation, 
for the missing views, all previous frames of this view are lost, so we mask both frames from $t-1$ and $t$. 
The next question is how to recover the features of these masked views.
In our task, the entire image is masked, we can not explore the relations within an image like the MAE style, but the relations between the six images instead.
As shown in Fig. \ref{fig:view}, 
six camera views share the overlapped regions,
which offer the distinct spatio-temporal clues,
i.e.,
the same object might appear in different views at different time steps.
Based on this observation,
we propose two types of MVR module to recover the features of masked views,
by using the features of rest views.

\noindent\textbf{Global MVR Module}.
In this design,
we use all the rest views as context to reconstruct the masked ones.
First,
we concatenate the features of all the views.
For the missing views,
we pad them with the shared and learned masked tokens which are randomly initialized, 
i.e., 
$\mathbf{V}_{mask}$. 
For the rest views,
we use their corresponding 2D features from encoder, 
i.e., 
$\mathbf{F}_{rest}$. 
Second,
we need to distinguish the location of these views to encode 3D relationships among them.
Hence,
we apply 3D position embedding \cite{liu2022petr} to encode 3D coordinates of different views,
i.e.,
$\mathbf{P}_{3D}$. 
Finally,
we add all the features with the corresponding position embedding,
and feed them into a feature decoder for reconstructing the masked features.
The decoder is composed of transformer blocks which are the same as MAE \cite{he2022masked},
except that the last layer outputs the feature tokens, 
instead of the raw pixels of the masked views.

\begin{small}
\begin{equation}
\mathbf{U}_{mask}= Decoder(\mathbcal{C}[\mathbf{V}_{mask},~\mathbf{F}_{rest}] + \mathbf{P}_{3D}),
\label{eq:Global}
\end{equation}
\end{small}
$\mathbf{U}_{mask}$ is the reconstructed features of the masked views, $\mathbcal{C}$ means we concat the tokens.

\noindent\textbf{Local MVR Module}.
In fact, 
a camera view is strongly correlated to its adjacent views,
instead of all the views.
As shown in Fig. \ref{fig:Overall Architecture}, 
the left part of Front view is relevant to the right part of Front\_Left view,
while
the right part of Front view is relevant to the left part of Front\_Right view.
Hence,
using all the views of Global MVR may introduce the noisy reconstruction,
with irrelevant surrounding scenes.
Based on this analysis,
we propose to design a local MVR module,
which uses the relevant context from the neighboring views to recover the missing views.
Specifically,
for the feature of a masked view,
we divide it into three parts, 
i.e.,
the left, middle, and right parts,
with a dividing portion ratio.
We investigate different portion ratios in our experiments.
(1) \textit{Left Part}:
As shown in Fig. \ref{fig:Overall Architecture},
for the left part of this masked view, 
we refer to the adjacent camera on its left view.
In particular,
we crop the right part of left-view features $\mathbf{F}_{left}(right)$ at $t-1$ and $t$ with the same ratio.
Then,
we use the mean of them as the left part of the masked view for both time stamps.
(2) \textit{Right Part}:
Similarly,
for the right part of this masked view, 
we refer to the adjacent camera on its right view.
In particular,
we crop the left part of right-view features $\mathbf{F}_{right}(left)$ at $t-1$ and $t$ with the same ratio.
Then,
we use the mean of them as the right part of the masked view for both time stamps.
(3) \textit{Middle Part}:
For the middle part of this masked view, 
there is no clue.
Hence,
we use the masked tokens just like Global MVR,
i.e., 
$\mathbf{V}_{mask}(mid)$.
As a result,
the 
masked view becomes the concatenation of these features,
\begin{small}
\begin{equation}
    \mathbf{V}_{mask}= \mathcal{C}[\mathbf{F}_{left}(right),
     ~\mathbf{V}_{mask}(mid),\\
     ~\mathbf{F}_{right}(left)].
    \label{eq:Local1}
\end{equation}
\end{small}
Multiple views can be randomly masked. 
If the adjacent views of a masked view are also missing,
there are no left and/or right clues for this view.
In this case,
we use the masked features $\mathbf{V}_{mask}(left)$ or/and $\mathbf{V}_{mask}(right)$ in the corresponding left or/and right parts.
Finally,
since $\mathbf{V}_{mask}$ is the 2D feature for a camera view,
we use 2D positional embedding $\mathbf{P}_{2D}$ by the sine-cosine version \cite{he2022masked},
and feed the sum of $\mathbf{V}_{mask}$ and $\mathbf{P}_{2D}$ into the decoder,
\begin{small}
\begin{equation}
\mathbf{U}_{mask}= Decoder(\mathbf{V}_{mask} + \mathbf{P}_{2D}),
\label{eq:Local2}
\end{equation}
\end{small}
where
$\mathbf{U}_{mask}$ is the reconstructed features of the masked views.
Additionally,
the decoder for Local MVR has the same transformer structure as the one for Global MVR.
But it only needs to process the tokens of one image size in Local MVR,
instead of all the tokens in the six images of Global MVR.
Hence,
the computation cost of Local MVR is less expensive than Global MVR.

\begin{table*}[t]
\small
\centering
    \begin{tabular}{l|l|cc|ccccc} 
    \hline
    \textbf{W/O Missing } & \textbf{Method} &  \textbf{NDS} $\uparrow$ & \textbf{mAP} $\uparrow$ & \textbf{mATE} $\downarrow$ & \textbf{mASE} $\downarrow$ & \textbf{mAOE} $\downarrow$ & \textbf{mAVE} $\downarrow$ & \textbf{mAAE} $\downarrow$\\
    \hline
    Standard & PETRv2 \cite{liu2023petrv2} & 0.4853 & 0.3977 & 0.7531	& 0.2693 & 0.4978	& 0.4310 &	0.1840 \\
    \hline
    \textbf{Missing} & \textbf{Method} & \textbf{NDS} $\uparrow$ & \textbf{mAP} $\uparrow$ & \textbf{mATE} $\downarrow$ & \textbf{mASE} $\downarrow$ & \textbf{mAOE} $\downarrow$ & \textbf{mAVE} $\downarrow$ & \textbf{mAAE} $\downarrow$\\
    \hline    
    \multirow{2}{*}{Front} & PETRv2 \cite{liu2023petrv2}   & 0.4238 & 0.3022 & 0.7757	& 0.2740 & 0.5311	& 0.5043 &	0.1883 \\
     & \textbf{Our M-BEV} (PETRv2)   & \textbf{0.4504} & \textbf{0.3263} & 0.7234	& 0.2736 & 0.4996	& 0.4449 &	0.1862 \\
    \hline    
    \multirow{2}{*}{Front\_Right} & PETRv2 \cite{liu2023petrv2} & 0.4363 & 0.3294 & 0.8444	& 0.2706 & 0.5333 & 0.4551 & 0.1808 \\
     & \textbf{Our M-BEV} (PETRv2)     & \textbf{0.4712} & \textbf{0.3666} & 0.7346	& 0.2704 & 0.5098 & 0.4214 & 0.1850 \\
    \hline    
    \multirow{2}{*}{Front\_Left} & PETRv2 \cite{liu2023petrv2} & 0.4405 & 0.3355 & 0.8195 & 0.2733	& 0.5216 & 0.4664 & 0.1912 \\
     & \textbf{Our M-BEV} (PETRv2)    & \textbf{0.4678} & \textbf{0.3628} & 0.7308 & 0.2740	& 0.5113 & 0.4298 & 0.1905 \\
    \hline    
    \multirow{2}{*}{Back} & PETRv2 \cite{liu2023petrv2} & 0.3616 & 0.2179 & 1.0176 & 0.2977 & 0.5618 & 0.4477 & 0.1726 \\
    & \textbf{Our M-BEV} (PETRv2)    & \textbf{0.4516} & \textbf{0.3206} & 0.7283 & 0.2688 & 0.4908 & 0.4237 & 0.1754 \\
    \hline    
    \multirow{2}{*}{Back\_Left} & PETRv2 \cite{liu2023petrv2} & 0.4568 & 0.3513 &  0.7910 & 0.2700 & 0.4903 & 0.4476 & 0.1895 \\
     & \textbf{Our M-BEV} (PETRv2)   & \textbf{0.4753} & \textbf{0.3694} & 0.7277 & 0.2694 & 0.4770 & 0.4291 & 0.1909 \\
    \hline    
    \multirow{2}{*}{Back\_Right} & PETRv2 \cite{liu2023petrv2} & 0.4556 & 0.3544 & 0.7892 & 0.2700 & 0.5157 & 0.4508 & 0.1902 \\
     & \textbf{Our M-BEV} (PETRv2)    & \textbf{0.4756} & \textbf{0.3730} & 0.7294 & 0.2711 & 0.4990 & 0.4211 & 0.1879 \\    
    \hline
    \end{tabular}

\caption{Performance comparison on PETRv2 \cite{liu2023petrv2} when losing each of six camera views. The effect of our M-BEV is impressive, e.g., for the absence of back view, our M-BEV achieves 10.3\% mAP improvement. 
}
\label{tab1new}      
\vspace{-0.3cm}     
\end{table*}

\begin{table*}[t]
\small
\centering
    \begin{tabular}{l|l|cc|ccccc} 
    \hline
    \textbf{W/O Missing } & \textbf{Method} &  \textbf{NDS} $\uparrow$ & \textbf{mAP} $\uparrow$ & \textbf{mATE} $\downarrow$ & \textbf{mASE} $\downarrow$ & \textbf{mAOE} $\downarrow$ & \textbf{mAVE} $\downarrow$ & \textbf{mAAE} $\downarrow$\\
    \hline
    Standard &  BEVStereo \cite{li2023bevstereo} & 0.4432 & 0.3439 & 0.6583	& 0.2823 & 0.5860	& 0.5287 &	0.2327\\
    \hline
    \textbf{Missing} & \textbf{Method} & \textbf{NDS} $\uparrow$ & \textbf{mAP} $\uparrow$ & \textbf{mATE} $\downarrow$ & \textbf{mASE} $\downarrow$ & \textbf{mAOE} $\downarrow$ & \textbf{mAVE} $\downarrow$ & \textbf{mAAE} $\downarrow$\\
    \hline    
    \multirow{2}{*}{Front} &  BEVStereo \cite{li2023bevstereo}   & 0.3901 & 0.2462 & 0.6799	& 0.2867 & 0.6192	& 0.5159 &	0.2283 \\
     & \textbf{Our M-BEV} (BEVStereo)   & 0.\textbf{4027} & \textbf{0.2667} & 0.6744	& 0.2866 & 0.6223	& 0.5103 &	0.2130 \\
    \hline    
    \multirow{2}{*}{Front\_Right} &  BEVStereo \cite{li2023bevstereo} & 0.4042 & 0.2832 & 0.6770	& 0.2841 & 0.6027 & 0.5811 & 0.2298 \\
     & \textbf{Our M-BEV} (BEVStereo)     & \textbf{0.4148} & \textbf{0.3004} & 0.6733	& 0.2845 & 0.6122 & 0.5791 & 0.2153 \\
    \hline    
    \multirow{2}{*}{Front\_Left} & BEVStereo \cite{li2023bevstereo} & 0.4039 & 0.2795 & 0.6720 & 0.2820	& 0.5960 & 0.5645 & 0.2429 \\
     & \textbf{Our M-BEV} (BEVStereo)   & \textbf{0.4126} & \textbf{0.2967} & 0.6725 & 0.2851	& 0.6134 & 0.5690 & 0.2269 \\
    \hline    
    \multirow{2}{*}{Back} &  BEVStereo \cite{li2023bevstereo}& 0.3894 & 0.2373 & 0.6761 & 0.2840 & 0.6071 & 0.5126 & 0.2126 \\
    & \textbf{Our M-BEV} (BEVStereo)    & \textbf{0.3949} & \textbf{0.2580} & 0.6703 & 0.2880 & 0.6196 & 0.5540 & 0.2087 \\
    \hline    
    \multirow{2}{*}{Back\_Left} &  BEVStereo \cite{li2023bevstereo} & 0.4132 & 0.2891 & 0.6753 & 0.2828 & 0.5687 & 0.5457 & 0.2408 \\
     & \textbf{Our M-BEV} (BEVStereo)  & \textbf{0.4149} & \textbf{0.2989} &  0.6762 & 0.2859 & 0.5966 & 0.5709 & 0.2259 \\
    \hline    
    \multirow{2}{*}{Back\_Right} &  BEVStereo \cite{li2023bevstereo} & 0.4081 & 0.2898 & 0.6813 & 0.2817 & 0.5945 & 0.5739 & 0.2367 \\
     & \textbf{Our M-BEV} (BEVStereo)   & \textbf{0.4157} & \textbf{0.3027} & 0.6719 & 0.2843 & 0.6052 & 0.5738 & 0.2209 \\ 
    \hline
    \end{tabular}

\caption{Performance comparison on BEVStereo \cite{li2023bevstereo} when losing each of six camera views. Overall, our M-BEV gives a comprehensive improvement for the baseline.}
\label{tab2new}  
\vspace{-0.4cm}     
\end{table*}

\subsection{Training and Testing}

\noindent\textbf{Training}.
After obtaining the reconstructed features of the masked views $\mathbf{U}_{mask}$ from MVR,
we compute the L2 loss between the original features $\mathbf{F}_{mask}$ and the reconstructed ones for self-supervised pretraining of our M-BEV, we freeze the encoder during pretraining.
\begin{small}
\begin{equation}
\mathcal{L}_{mvr}=\Vert \mathbf{F}_{mask}-\mathbf{U}_{mask} \Vert_{2}.
\label{mvrerror}
\end{equation}
\end{small}
However, only feature supervision is not enough for perception,
we further fine-tune our M-BEV with 3D detection supervision end to end,
\begin{small}
\begin{equation}
\mathcal{L}_{total}=\mathcal{L}_{det}+\alpha\mathcal{L}_{mvr},
\label{totalerror}
\end{equation}
\end{small}
where
$\mathcal{L}_{det}$ is the 3D detection loss which consists of focal loss \cite{lin2017focal} for object classification and L1 loss for 3D bounding box regression.
Since our goal is to boost detection in case of emergency,
the detection loss should be the major loss.
In this case,
we set a weight coefficient $\alpha=0.05$ for the reconstruction loss in the fine-tuning.
Moreover,
we randomly select different number of camera views in different training epochs,
e.g.,
one view can be randomly picked for masking in the previous epoch,
while
four views can be randomly picked for masking in the current epoch.
This is different from the training style of MAE \cite{he2022masked},
where the same masking proportion is adopted in all the epochs.
The main reason is that
our goal is to tackle all the missing view cases which can possibly happen.
Hence,
the training procedure should cover various missing views by random masking in different epochs.
This design has not been attempted in previous research on mask modeling.

\noindent\textbf{Testing}.
In the testing process,
we preserve the view decoder in MVR to predict the features of missing views.
This design can effectively tackle the camera failure emergency in the driving procedure.
Additionally,
for the regular case without missing views,
our M-BEV still works well,
where
we can simply ignore the trained MVR and feed the features of encoder into 3D detection decoder.
In fact,
our trained M-BEV is better for the regular case,
compared to the baseline model (without view masking and recovering).
This is mainly because that
training with MVR can generalize our M-BEV model for various hard masking cases.
In our experiment,
we have validated this conclusion (Table. \ref{tab5}). 

\begin{figure*}[t]
\centering

\includegraphics[width=0.85\textwidth]{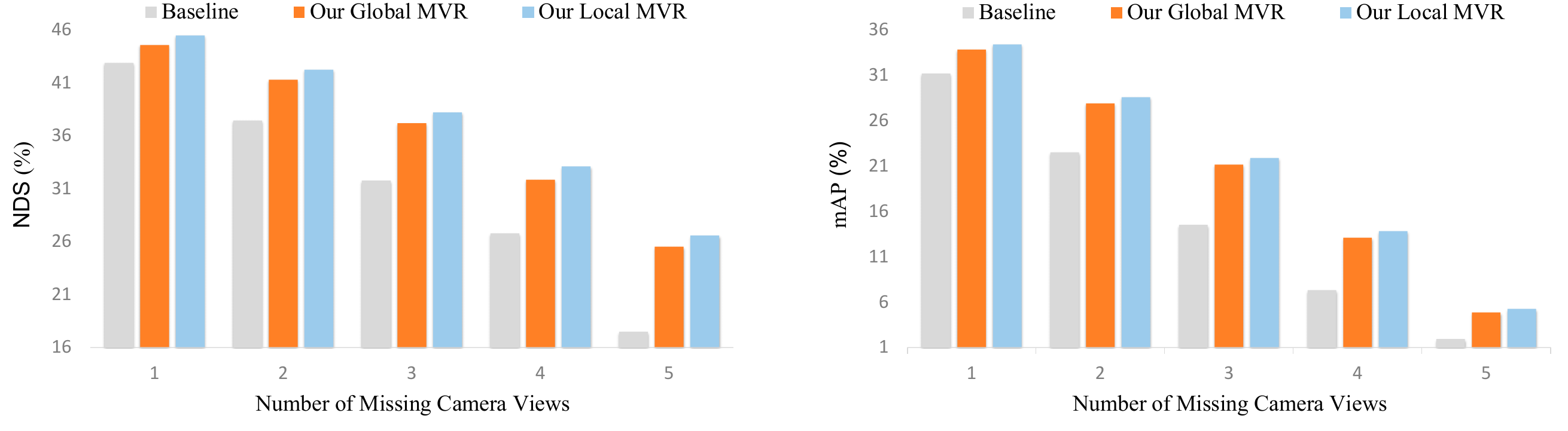} 
\caption{Performance of baseline, our Global MVR and Local MVR settings for different number of missing cameras.} 
\label{fig:line2}
\vspace{-0.4cm}     
\end{figure*}

\section{Experiments}

\subsection{Datasets and Metrics}
We conduct our experiments on the popular 
NuScenes dataset \cite{caesar2020nuscenes}. 
NuScenes is a large-scale benchmark for autonomous driving, where the data is collected from 1000 real driving scenes with around 20 seconds duration. The scenes are divided: 700 of them for training, and 150 each for validation and testing.
Two keyframes are annotated per second for the whole 1.4M 3D bounding boxes.
We report the officially used metrics of 3D object detection in BEV-based research\cite{caesar2020nuscenes,lang2019pointpillars,wang2022detr3d} ,
i.e.,
mean Average Precision(mAP) and five True Positive metrics, 
including 
mean Average Translation Error (mATE), 
mean Average Scale Error (mASE), 
mean Average Orientation Error(mAOE), 
mean Average Velocity Error(mAVE), 
mean Average Attribute Error(mAAE), 
where the lower value is better. 
Besides, 
the NuScenes Detection Score (NDS) comprehensively reflects these metrics, 
and it is the most concerned metric for performance evaluation. 

\subsection{Implementation Details} 
Driving scenes in the real world are often complex, we choose a challenging setting to mimic real situation, that is, we randomly discard images of the corresponding views using our RVM module, all previous frames for the view are also missing, so we can evaluate how other views help for the reconstruction of the missing ones.
For the baseline models, we follow the official implementation on open-sourced code bases. 
For PETRv2\cite{liu2023petrv2}, we use images of 320×800 resolution as input, and the visual backbone is pre-trained VoVNet-99\cite{lee2020centermask}.
For BEVStereo\cite{li2023bevstereo}, the model is obtained by the official code on GitHub with ResNet-50\cite{he2016deep} as visual encoder and uses 256x704 input resolution. 
The models are trained with official settings and get comparable performance with the official report.
And for inference, we do the evaluation on all possible situations. 
The MVR module is fine-tuned for 48 epochs, the learning rate is set to $2.0 \times 10^{-4}$. 
The transformer layer of decoder is four, and the hidden dimension is 512.
We use 8 A5000 GPUs for all experiments.
No test-time augmentation methods are used during inference. 
\vspace{-0.4cm}

\subsection{SOTA Comparison}
As mentioned before,
we apply M-BEV paradigm on two recent state-of-the-art approaches,
PETRv2\cite{liu2023petrv2} and BEVStereo\cite{li2023bevstereo},
where
we insert our local MVR after the visual encoder in these models.
We compare our M-BEV paradigm with the original BEV paradigm on these models.
Moreover,
to explicitly evaluate the effectiveness,
we investigate the result for missing each of the six camera views.
As shown in Table \ref{tab1new} and Table \ref{tab2new},
our M-BEV comprehensively improves the performance,
compared to the original model.
For example,
M-BEV is remarkable for the back-view camera failure, with a 10.3\% mAP growth on the PETRv2 baseline.
The recovered result for the back\_right view is even comparable with the setting without missing views.
All these prove the effectiveness of our design.

\subsection{Ablation Study}
\begin{table*}[t]
\small
\centering
    \begin{tabular}{l|c|cc|ccccc} 
    \hline
    \textbf{Missing} & \textbf{Our M-BEV} & \textbf{NDS} $\uparrow$ & \textbf{mAP} $\uparrow$ & \textbf{mATE} $\downarrow$ & \textbf{mASE} $\downarrow$ & \textbf{mAOE} $\downarrow$ & \textbf{mAVE} $\downarrow$ & \textbf{mAAE} $\downarrow$\\
    \hline    
    \multirow{2}{*}{Front} &  Global MVR   & 0.4361 & 0.3160 & 0.7524	& 0.2725 & 0.5489	& 0.4584 &	0.1873 \\
     & Local MVR     & \textbf{0.4504} & \textbf{0.3263} & 0.7234	& 0.2736 & 0.4996	& 0.4449 &	0.1862 \\
    \hline    
    \multirow{2}{*}{Front\_Right} &  Global MVR & 0.4588 & 0.3592 & 0.7622	& 0.2684 & 0.5550 & 0.4371 & 0.1852 \\
     & Local MVR   & \textbf{0.4712} & \textbf{0.3666} & 0.7346	& 0.2704 & 0.5098 & 0.4214 & 0.1850 \\
    \hline    
    \multirow{2}{*}{Front\_Left} & Global MVR & 0.4574 & 0.3570 & 0.7635 & 0.2727	& 0.5410 & 0.4410 & 0.1928 \\
     & Local MVR   & \textbf{0.4678} & \textbf{0.3628} & 0.7308 & 0.2740	& 0.5113 & 0.4298 & 0.1905 \\
    \hline    
    \multirow{2}{*}{Back} &  Global MVR & 0.4390 & 0.3146 & 0.7642 & 0.2679 & 0.5340 & 0.4400 & 0.1773 \\
    & Local MVR  & \textbf{0.4516} & \textbf{0.3206} & 0.7283 & 0.2688 & 0.4908 & 0.4237 & 0.1754 \\
    \hline    
    \multirow{2}{*}{Back\_Left} &  Global MVR & 0.4632 & 0.3633 &  0.7645 & 0.2697 & 0.5214 & 0.4350 & 0.1944 \\
     & Local MVR  & \textbf{0.4753} & \textbf{0.3694} & 0.7277 & 0.2694 & 0.4770 & 0.4291 & 0.1909 \\
    \hline    
    \multirow{2}{*}{Back\_Right} &  Global MVR & 0.4629 & 0.3671 & 0.7638 & 0.2694 & 0.5504 & 0.4333 & 0.1895 \\
     & Local MVR  & \textbf{0.4756} & \textbf{0.3730} & 0.7294 & 0.2711 & 0.4990 & 0.4211 & 0.1879 \\
    \hline
    \end{tabular}

\caption{The performance after reconstruction of two MVR module variants. Local MVR is a better choice by exploiting the distinct context from adjacent camera views. }
\label{tab3new}  
\vspace{-0.5cm}
\end{table*}

\begin{table*}[t]
\small
\centering

    \begin{tabular}{cccc|cc|ccccc} 
    \hline
       \textbf{MVR} 
    &  \textbf{FineTune} 
    & \textbf{PE} 
    & \textbf{MaskRatio} 
    & \textbf{NDS} $\uparrow$ & \textbf{mAP} $\uparrow$ & \textbf{mATE} $\downarrow$ & \textbf{mASE} $\downarrow$ & \textbf{mAOE} $\downarrow$ & \textbf{mAVE} $\downarrow$ & \textbf{mAAE} $\downarrow$\\
    \hline
    
      \XSolidBrush 
    & \XSolidBrush 
    & \XSolidBrush 
    & \XSolidBrush  
    & 0.4285 & 0.3117  &  0.8279 & 0.2730 & 0.5228 & 0.4638  & 0.1865  \\

      \XSolidBrush 
    & \Checkmark
    & \XSolidBrush 
    & \XSolidBrush
    & 0.4466 & 0.3371 & 0.7752  & 0.2741  & 0.5256  &  0.4469 & 0.1975  \\

     \Checkmark
    & \XSolidBrush 
    & \Checkmark
    & 76\%
    & 0.4319 & 0.3130 &  0.8196 & 0.2721  & 0.5181 & 0.4545 & 0.1852    \\

     \Checkmark 
    & \Checkmark
    & \XSolidBrush
    & 76\%
    & 0.4552 & 0.3477 &  0.7556 & 0.2715 & 0.5190 & 0.4376 & 0.1870         \\

      \Checkmark
    & \Checkmark
    & \Checkmark
    & 76\%
    & \textbf{0.4580} & \textbf{0.3499} & 0.7544 & 0.2712 & 0.5186 & 0.4368 & 0.1878   \\

    \hline

      \Checkmark
    & \Checkmark
    & \Checkmark
    & 60\% & 0.4576 & 0.3495 & 0.7538 &  0.2710 & 0.5171 & 0.4402 & 0.1885 \\

      \Checkmark
    & \Checkmark
    & \Checkmark
    & 64\% & 0.4566 & 0.3474 & 0.7529 & 0.2720 & 0.5185 & 0.4378 & 0.1894 \\

      \Checkmark
    & \Checkmark
    & \Checkmark
    & 68\% & 0.4570 & 0.3491 & 0.7596 & 0.2714 & 0.5246 & 0.4302 & 0.1892 \\

      \Checkmark
    & \Checkmark
    & \Checkmark
    & 72\% & 0.4579 & 0.3485 & 0.7540 & 0.2721 & 0.5126 & 0.4352 & 0.1895 \\

      \Checkmark
    & \Checkmark
    & \Checkmark
    & 80\%  & 0.4560 & 0.3478 & 0.7578 & 0.2716 & 0.5249 & 0.4330 & 0.1916 \\
    \hline    
    
    \end{tabular}

\caption{The ablation of MVR module, fine-tuning, position embedding and masking ratio for M-BEV. 
All models are trained with same schedule and hyper-parameters. 
We randomly drop one camera and calculate the average metrics for inference.}
\label{tab4}
\vspace{-0.3cm}     
\end{table*}

\textbf{Global MVR \textit{v.s.} Local MVR}.
We first ablate the effect of Global MVR and Local MVR,
where we use the PETRv2 baseline due to its good performance.
To ensure fairness, 
both two MVR variants are trained with the same schedule when each camera view loses, 
and all other hyper-parameters keep the same.
The results are shown in Table. \ref{tab3new}, Local MVR method outperforms Global MVR in all metrics.
This superiority may be attributed to the fact that, 
Local MVR exploits the distinct spatio-temporal context across adjacent cameras,
instead of using all the views which may contain noise.

\noindent\textbf{Number of Missing Camera Views}.
It is natural to evaluate the robustness if more than one camera is lost in real-world scenarios. 
Note that,
there are several possible missing view choices for each setting,
e.g.,
there are 6/15/20/15/6 choices for missing 1/2/3/4/5 views.
Hence,
we compute the NDS and mAP metrics for each choice and average them as the final NDS and mAP metrics.
As mentioned in the training section,
we train a single model to handle all these situations to show our robustness.
As shown in Fig. \ref{fig:line2},
our method outperforms the baseline PETRv2 for all the missing cases,
especially when the number of missing views increases.
It clearly shows the robustness of our M-BEV.
Moreover, Local MVR is consistently better than Global MVR,
based on the exploration of distinct contexts across adjacent views.

\begin{figure*}[t]
\small
\centering
\setlength{\abovecaptionskip}{0.cm}
\includegraphics[width=0.9\textwidth]{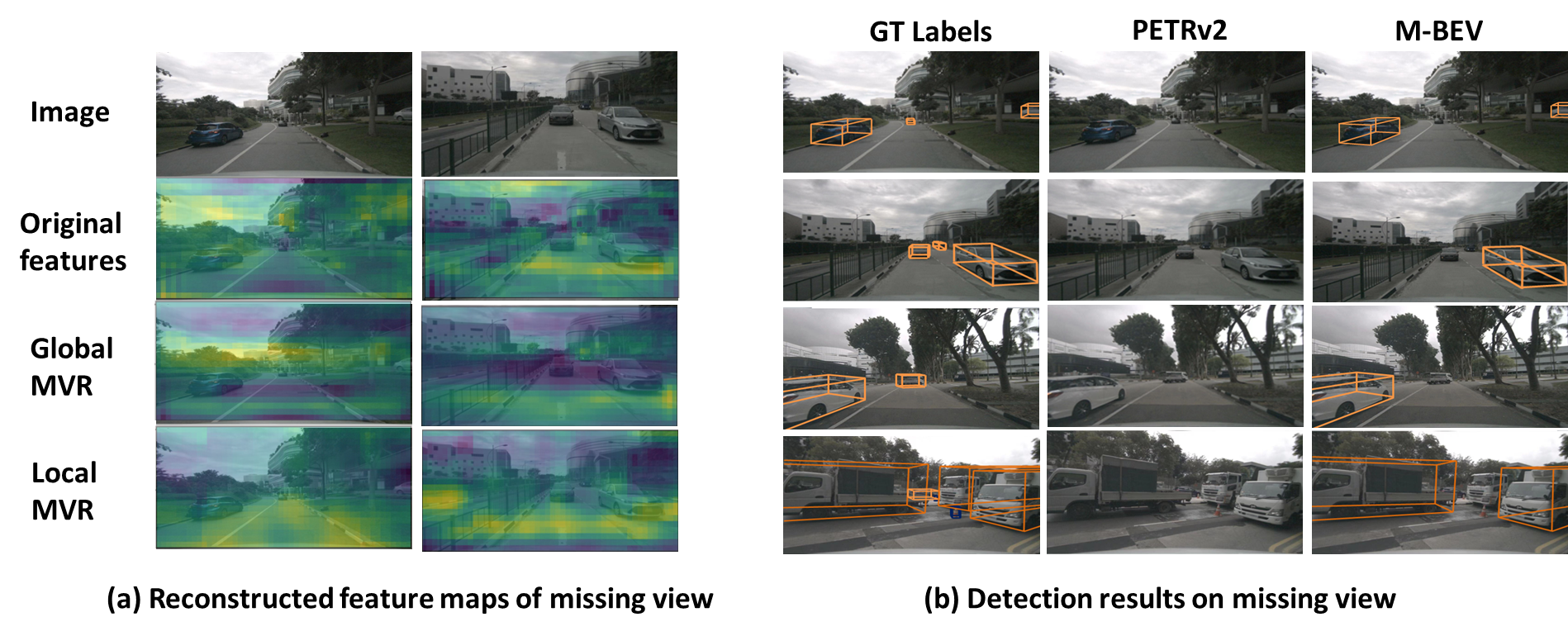} 
\caption{Visualization for feature maps and detection results. M-BEV reconstructed
features could be a rough substitute of the original feature, and could help for the detection of the vehicles on the left and right sides of the missing view.  } 
\label{fig:vis}
\vspace{-0.3cm}     
\end{figure*}

\noindent\textbf{MVR Designs}.
To verify the contributions of the strategies in our proposed MVR module, we conduct ablation experiments on several settings which will be explained in detail below.
All ablations are conducted with the Local MVR method which performs better.
In Table\ref{tab4}, we ablate the MVR module employed in the model, fine-tuning process, the position embedding(PE), and the masking ratio in sequence to validate how they contribute to the final results, the results show the average values under one random camera failure.
Note that the best performance is obtained with all the strategies used and 76\% masking ratio, put in bold in Table\ref{tab4}.
(1)First, we ablate the MVR module which is the core of our M-BEV approach. 
To verify its effectiveness, we remove our proposed MVR module and perform the same fine-tuning using RVM module.
The final results show a drop in NDS and mAP by 1.14\% and 1.28\% respectively compared to the best model with MVR, 
indicating that our reconstructed features could offer extra information beyond the original model. 
(2)Then we explore the influence of fine-tuning process which aims to alleviate the domain gap between reconstruction task and detection task.
The model without fine-tuning shows a drop in NDS and mAP by 2.61\% and 3.69\%, which strongly confirms the significance of fine-tuning with our RVM module.
(3)Next, as the composition of input for MVR decoder, PE is added to the 2D feature tokens for better localization. 
The ablation shows the influence of PE which indeed makes a progress.
The model with PE have an improvement of 0.28\% NDS and 0.22\% mAP on average compared to model without it.
(4)Finally, we ablate the masking ratio.
The masking ratio depends on the dividing portion we use from the neighboring overlaps, the left and right views could both offer prompts for the missing camera, so we only need to mask the middle part.
A proper ratio close to the real situation is also crucial for good reconstruction.
We have evaluated different masking ratios from 60\% to 80\%, 
the optimal masking ratio is 76\% for both NDS and mAP.

\noindent\textbf{Segmentation}.
We also evaluate our M-BEV for map segmentation tasks on nuScenes with PETRv2\cite{liu2023petrv2} baseline, where only the prediction head needs to be changed. 
When missing one camera view, the IoU scores of Drive, Lane and Vehicle drop form 79.5\%, 46.2\% and 49.9\% to 76.6\%, 41.1\% and 43.5\% respectively on average, while with our MVR, the IoU scores are 78.2\%, 45.1\% and 45.6\%, much better than that of original model.

\noindent\textbf{Generalization vs. Computation}.
Finally,
we evaluate the generalization and computation cost of M-BEV. 
As shown in the Table. \ref{tab5}, 
for the regular cases without missing views, 
we can directly deactivate the MVR module after training,
the GFLOPS keep the same as the baseline.
However,
the model co-trained with MVR performs better than the baseline,
showing that, 
M-BEV paradigm can generalize the learning capacity for 3D object detection,
by masking view modeling.
For the camera failure case (one view missing setting), 
local MVR has better performance with little extra computation cost, and it only takes about 6ms for the local MVR to response for once detection, which is negligible.
All these prove its potential for practical application.

\begin{table}[t]
\small
\resizebox{\columnwidth}{!}{    
\begin{tabular}{l|cccc}   
\hline
\textbf{W/O Missing} & \textbf{NDS}$\uparrow$ &  \textbf{mAP}$\uparrow$ & \textbf{GFLOPS}$\downarrow$ & \textbf{Time(ms)}$\downarrow$\\ 
\hline
PETRv2 Baseline & 0.4853  & 0.3977 &  1047 & -- \\    
Our Global MVR  & 0.4872  & 0.4007  & 1047 & +0.0  \\    
Our Local MVR & \textbf{0.4898} & \textbf{0.4039}  &  1047 & +0.0 \\       
\hline 
\textbf{Missing} & \textbf{NDS}$\uparrow$ &  \textbf{mAP}$\uparrow$ & \textbf{GFLOPS}$\downarrow$ & \textbf{Time(ms)}$\downarrow$\\ 
\hline          
PETRv2 Baseline & 0.4285 & 0.3117 & 1047 & --  \\    
Our Global MVR  & 0.4534  & 0.3467  & 1088  &  +1.94\\    
Our Local MVR & \textbf{0.4580}  & \textbf{0.3499}  &  1053 & +0.61 \\       
\hline 
\end{tabular}
} 
\setlength{\abovecaptionskip}{0.cm}
\caption{Generalization vs. Computation. 
The model trained with our MVR performs better the original baseline under both no-missing and missing settings, while requiring only little extra GFLOPS for computation.
}
\label{tab5}
\vspace{-0.4cm}      
\end{table}

\vspace{-0.3cm}
\subsection{Visualization}

In Fig. \ref{fig:vis}, we give some examples of the reconstructed features of missing cameras and the detection results of the missing view.
As shown in Fig. \ref{fig:vis}, our M-BEV reconstructed features could be a rough substitute of the original features, from which we can see the outline of the road and the major targets.
For the detection results, our M-BEV is helpful for detection of the vehicles on left and right sides of the missing view. 
For example, due to the missing of back view, PETRv2 model can't detect any object in the view, but with our reconstruction, the vehicles near the overlap regions could be detected,
while it's still hard to detect the small and far objects in the middle part, which may need further exploration.

\vspace{-0.3cm}

\section{Conclusion and Future Work}
Recent research has primarily focused on improving detection performance by designing different detectors for BEV features and incorporating depth supervision.
Our work focuses on improving the robustness of these models,
which is essential for ensuring driving safety.
In this paper, we put forward a novel reconstruction architecture to address the emergence of camera crashes.
To compensate for the lost information of missing camera views, 
we design a distinct MVR module that leverages the related tokens from neighboring cameras.
The reconstructed image features are capable of boosting the detection results,
compared to the original models.
Furthermore, M-BEV has great generalization ability and requires little extra computation.
Extensive experiments verify the effectiveness of our M-BEV, which could be widely applied as a plug-and-play module to enhance the robustness of 3D object detection models.

\noindent\textbf{Future work.}
A comprehensive system with multiple tasks needs extensive experiment verification, our M-BEV needs more future exploration with various benchmarks
to achieve a framework to unify perception, prediction and planning tasks for robust auto-driving, 

\section{Acknowledgements}
This work was supported by the National Key R\&D Program of China(NO.2022ZD0160505),  and the Joint Lab of CAS-HK.

\bibliography{CameraReady/LaTeX/myref}
\end{document}